\documentclass{article}

\usepackage[preprint]{neurips_2026}

\usepackage{hyperref}
\usepackage{url}
\usepackage{booktabs}
\usepackage{amsfonts}
\usepackage{amsmath}
\usepackage{amssymb}
\usepackage{nicefrac}
\usepackage{microtype}
\usepackage{xcolor}
\usepackage{graphicx}
\usepackage{subcaption}
\usepackage{multirow}
\usepackage{adjustbox}
\usepackage{enumitem}

\title{PromptEcho: Annotation-Free Reward from Vision-Language Models for Text-to-Image Reinforcement Learning}

\author{%
  Jinlong Liu$^{*1}$, Wanggui He$^{*1}$, Peng Zhang$^{2}$, Mushui Liu$^{1}$, Hao Jiang$^{\dagger 1}$, Pipei Huang$^{1}$ \\
  $^{1}$Alibaba Group \quad $^{2}$Zhejiang University \\
  $^{*}$Equal contribution \quad $^{\dagger}$Corresponding author
}

\begin{document}

\maketitle

\begin{abstract}
Reinforcement learning (RL) can improve the prompt following capability of text-to-image (T2I) models, yet obtaining high-quality reward signals remains challenging: CLIP Score is too coarse-grained, while VLM-based reward models (e.g., RewardDance) require costly human-annotated preference data and additional fine-tuning. We propose PromptEcho, a reward construction method that requires \emph{no} annotation and \emph{no} reward model training. Given a generated image and a guiding query, PromptEcho computes the token-level cross-entropy loss of a frozen VLM with the original prompt as the label, directly extracting the image-text alignment knowledge encoded during VLM pretraining. The reward is deterministic, computationally efficient, and improves automatically as stronger open-source VLMs become available. For evaluation, we develop DenseAlignBench, a benchmark of concept-rich dense captions for rigorously testing prompt following capability. Experimental results on two state-of-the-art T2I models (Z-Image and QwenImage-2512) demonstrate that PromptEcho achieves substantial improvements on DenseAlignBench (+26.8pp / +16.2pp net win rate), along with consistent gains on GenEval, DPG-Bench, and TIIFBench without any task-specific training. Ablation studies confirm that PromptEcho comprehensively outperforms inference-based scoring with the same VLM, and that reward quality scales with VLM size. We have open-sourced the trained models and the DenseAlignBench (\url{https://github.com/roooobotx/prompt_echo}).
\end{abstract}

\section{Introduction}
\label{sec:intro}

Text-to-image (T2I) generation has advanced rapidly from GANs to Diffusion Models and Flow Matching architectures, yet prompt following---the precise semantic alignment between generated images and input text---remains a core challenge. Complex descriptions involving multiple objects, attribute binding, spatial relationships, or precise counting still cause frequent omissions and errors in state-of-the-art models.

Reinforcement learning (RL) has proven effective for improving prompt following. DDPO~\citep{black2024ddpo} and DPOK~\citep{fan2024dpok} model diffusion denoising as a Markov Decision Process and optimize external reward signals via policy gradients; Flow-GRPO~\citep{liu2025flowgrpo} and AWM~\citep{awm2025} extend this framework to Flow Matching models. However, the quality of the reward signal remains the bottleneck. CLIP Score~\citep{hessel2021clipscore} is too coarse-grained to capture attribute binding, spatial relations, or counting. Reward models trained on human preferences---PickScore~\citep{kirstain2023pickapic}, ImageReward~\citep{xu2023imagereward}, HPS~\citep{wu2023hpsv2}---improve over CLIP but are constrained by small model sizes and limited annotation coverage. Recent VLM-based reward models~\citep{xu2024visionreward,wang2025unifiedreward,rewarddance2025} raise the capability ceiling by fine-tuning VLMs on annotated data, yet they still require costly annotation and additional training. An alternative is to directly prompt VLMs for zero-shot inference scoring (which we term InferScore), but this suffers from hallucinations and sampling randomness (Section~\ref{sec:ablation}).

\paragraph{PromptEcho.}
We propose PromptEcho, a reward construction paradigm that sidesteps both annotation and reward model training. Rather than having the VLM \emph{reason and judge} image-text alignment, we directly extract the alignment knowledge already encoded in its pretrained weights. Specifically, given an image $x$ generated from prompt $c$, we feed $x$ and a guiding query (e.g., ``Describe this image in detail.'') into a frozen VLM, then use $c$ as the label to compute the token-level cross-entropy loss; the negative of this loss serves as the reward. Intuitively, if the image faithfully depicts the prompt, the VLM can ``recite'' $c$ with high probability after seeing the image---the image \emph{echoes} the prompt through the VLM. Because this loss is identical to the VLM's pretraining objective, PromptEcho maximally preserves the image-text alignment knowledge the VLM acquired from massive data. The reward is fully deterministic (no autoregressive sampling), computationally efficient (a single forward pass), and automatically improves as stronger open-source VLMs become available---requiring no re-annotation or retraining.

\paragraph{Contributions.}
\begin{itemize}[leftmargin=1.5em, itemsep=2pt, topsep=2pt]
\item We propose PromptEcho, the first method to directly use VLM cross-entropy loss as a reward signal for T2I reinforcement learning, opening a fundamentally new path for reward construction.
\item PromptEcho requires zero annotation and zero reward model training. By directly leveraging pretrained open-source VLMs, it achieves high-quality, deterministic reward signals whose quality automatically scales with VLM capability.
\item Extensive experiments on two state-of-the-art T2I models (Z-Image and QwenImage-2512) demonstrate substantial win rate improvements on DenseAlignBench (+26.8pp / +16.2pp), consistent gains on public benchmarks (GenEval, DPG-Bench, TIIFBench) without task-specific training, and successful application to text rendering---validating PromptEcho as a general-purpose reward paradigm.
\end{itemize}

\section{Related Work}
\label{sec:related}

\subsection{Reinforcement Learning for Text-to-Image Models}

Research on applying reinforcement learning to optimize diffusion models has progressed rapidly in recent years. DDPO (Denoising Diffusion Policy Optimization)~\citep{black2024ddpo} was the first to formalize the denoising process of diffusion models as a multi-step Markov Decision Process (MDP), treating each denoising step as an action and optimizing external rewards via policy gradient methods. DPOK~\citep{fan2024dpok} builds upon DDPO by introducing KL divergence regularization, constraining the optimized policy from deviating too far from the original pretrained distribution, thereby maintaining image quality and diversity while improving target attributes.

With the rise of Flow Matching models, RL optimization methods have been extended accordingly. Flow-GRPO~\citep{liu2025flowgrpo} adapts the Group Relative Policy Optimization (GRPO) algorithm to the continuous Flow Matching framework, performing policy updates through group-relative advantage estimation. However, the exploration noise introduced during Flow-GRPO training can cause sampled trajectories to deviate from the normal distribution, leading to image quality degradation. AWM (Advantage Weighted Matching)~\citep{awm2025} proposes a gentler optimization strategy: optimizing directly in the ODE prediction space via advantage-weighted MSE loss, avoiding the quality degradation caused by SDE exploration. Our experiments adopt AWM as the RL training framework.

\subsection{Reward Signal Sources}

Reward signals in T2I RL optimization can be categorized by their source and construction method.

\paragraph{Traditional reward metrics.} CLIP Score~\citep{hessel2021clipscore} relies on coarse-grained vector similarity and cannot capture complex semantic alignment. PickScore~\citep{kirstain2023pickapic}, ImageReward~\citep{xu2023imagereward}, HPS~\citep{wu2023hpsv2}, and others train dedicated reward models on human-annotated preference data, but are limited by small model sizes and finite training data, struggling to provide effective optimization signals for frontier T2I models. Methods such as Diffusion-DPO~\citep{wallace2024diffusiondpo} and Diffusion-RPO~\citep{gu2024diffusionrpo} bypass explicit reward models but still depend on human-annotated preference pairs.

\paragraph{VLM-based reward construction.} Recent works upgrade reward models to the VLM level, following two main paradigms: (1) Regression-based---VisionReward~\citep{xu2024visionreward} fine-tunes a VLM to answer fine-grained binary (yes/no) questions about image quality, then aggregates the per-question probabilities into a scalar reward via learned linear weights; (2) Generative---RewardDance~\citep{rewarddance2025} and UnifiedReward~\citep{wang2025unifiedreward} cast scoring as a VLM next-token prediction task, fine-tuning the VLM on human-annotated preference or instruction data to directly generate evaluation scores. While these methods raise the capability ceiling of reward models, all require human-annotated data and additional VLM fine-tuning. An alternative is to directly use VLMs for zero-shot inference scoring (InferScore), but this suffers from hallucinations and sampling randomness, producing unstable reward signals (we experimentally validate this in Section~\ref{sec:ablation}).

\paragraph{Positioning of PromptEcho.} Unlike all the above VLM-based reward methods, PromptEcho requires no annotation data and no reward model training, directly extracting reward signals from a frozen VLM through deterministic forward passes, with reward quality that improves as open-source VLMs advance.

\subsection{Vision-Language Models}

The rapid development of open-source VLMs provides a solid technical foundation for PromptEcho. Representative VLM families include LLaVA~\citep{liu2024llava}, which pioneered the visual instruction tuning paradigm; the InternVL series~\citep{chen2024internvl}, which scales up vision encoders and achieves strong multimodal reasoning; and the Qwen-VL series~\citep{bai2023qwenvl}, which has continuously evolved from Qwen-VL through Qwen2-VL and Qwen2.5-VL to Qwen3-VL, with iterative improvements in model scale, training data, and multimodal understanding capabilities. This paper adopts Qwen3-VL-32B as the source VLM for reward signals, owing to its strong image-text alignment capability at the 32B parameter scale. Importantly, PromptEcho is not tied to any specific VLM---when a more powerful open-source VLM is released, one simply replaces the VLM to obtain higher-quality reward signals, without re-collecting annotation data or retraining a reward model.

\section{Method}
\label{sec:method}

\subsection{PromptEcho Reward}
\label{sec:promptecho}

Let $G_\theta$ be a T2I model with parameters $\theta$ and $c$ the input text prompt. The goal of T2I RL is to maximize the expected reward: $\theta^* = \arg\max_\theta \, \mathbb{E}_{c \sim \mathcal{D}, x \sim G_\theta(c)} [ R(x, c) ]$, where $\mathcal{D}$ is the prompt distribution. We propose a new reward function $R_{\text{PromptEcho}}$ computed from a frozen VLM $\mathcal{M}$ as follows:

\paragraph{Input Construction.}
The reward computation takes three inputs: (1)~the generated image $x = G_\theta(c)$; (2)~a fixed guiding query $q$ (e.g., ``Describe this image in detail.'') that instructs the VLM to produce a description; and (3)~the original prompt $c$, which serves as the ground-truth label for cross-entropy computation.

\paragraph{PromptEcho Reward Definition.}
Given the VLM's output logits, we compute the cross-entropy loss only over the token positions corresponding to prompt $c$ and negate it to obtain the reward:
\begin{equation}
\label{eq:reward}
R_{\text{PromptEcho}}(x, c) = -\mathcal{L}_{\text{CE}}(c \mid x, q) = -\frac{1}{|c|}\sum_{t=1}^{|c|} \left[-\log p_{\mathcal{M}}(c_t \mid c_{<t}, x, q)\right]
\end{equation}
where $|c|$ is the number of tokens in prompt $c$, and $p_{\mathcal{M}}(c_t \mid c_{<t}, x, q)$ is the probability that VLM $\mathcal{M}$ predicts the $t$-th token $c_t$ given image $x$, query $q$, and prefix tokens $c_{<t}$.

\paragraph{Intuition.} If image $x$ faithfully depicts prompt $c$ (e.g., correctly rendering all objects, attributes, and spatial relationships), the VLM should ``recite'' each token of $c$ with high probability after observing the image---the image \emph{echoes} the prompt---yielding a high reward. Conversely, omissions or semantic deviations in the image lower the token-level likelihood and thus the reward. This formulation works because the reward is exactly the VLM's pretraining objective (token-level cross-entropy), directly tapping into the image-text alignment knowledge acquired from massive data---without any additional conversion layers, regression heads, or fine-tuning. We analyze this mechanism further in Section~\ref{sec:understanding}.

\begin{figure}[t]
  \centering
  \includegraphics[width=\linewidth]{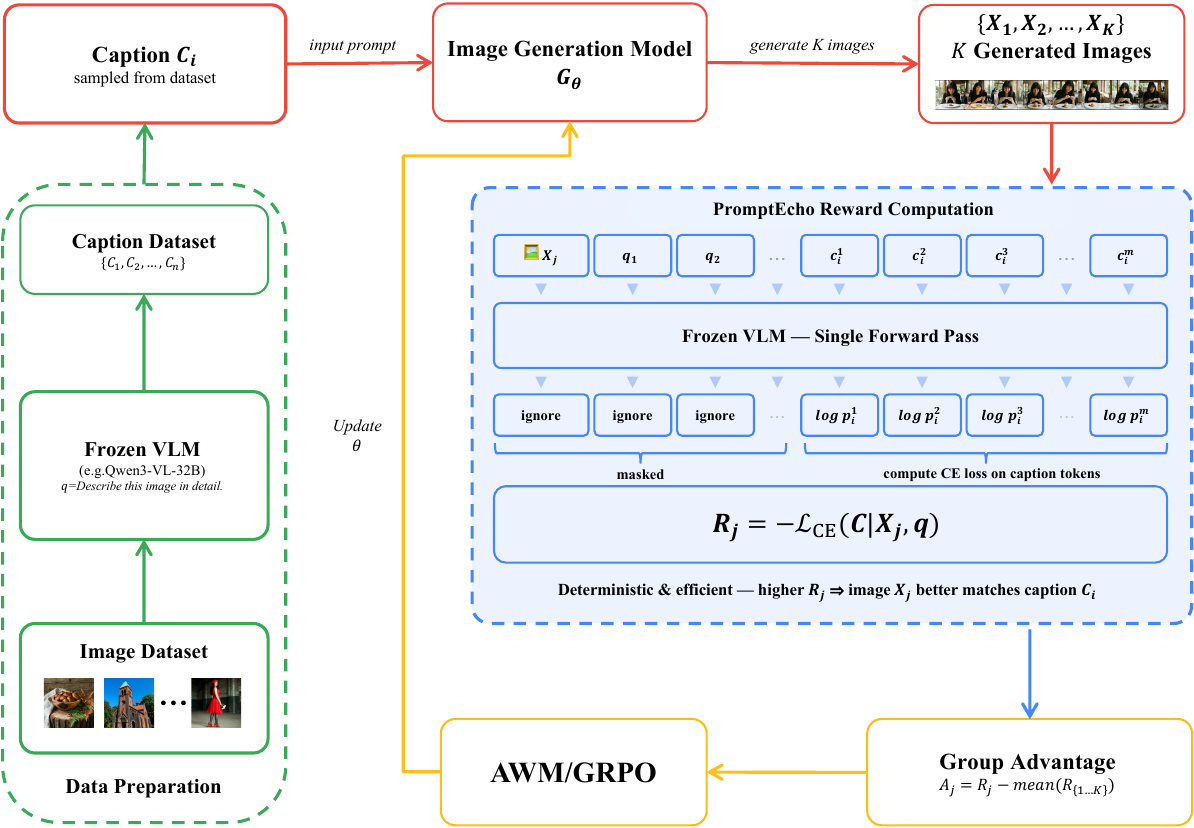}
  \caption{Overview of the PromptEcho training pipeline. \textbf{(Left, dashed green)} \textit{Data Preparation}: a frozen VLM captions images from a source dataset to build the caption corpus $\{c_1, \ldots, c_N\}$. \textbf{(Main loop)} At each iteration, a caption $c_i$ is sampled and fed to the T2I model $G_\theta$, which generates $K$ images. Each image $x_j$ enters the \textbf{PromptEcho Reward Computation} (blue box): the image, a fixed query $q$, and the original caption $c$ are concatenated as input tokens to the same frozen VLM; a single teacher-forcing forward pass produces logits over all positions; the cross-entropy loss is computed only on caption token positions, yielding the reward $R_j = -\text{CE}(c \mid x_j, q)$. Group advantage $\hat{A}_j = R_j - \text{mean}(R_{1:K})$ is then used by AWM/GRPO to update $G_\theta$ via policy gradient. The same frozen VLM (e.g.\ Qwen3-VL-32B) is shared across data preparation and reward computation, requiring no fine-tuning or human annotation.}
  \label{fig:pipeline}
\end{figure}

\subsection{RL Training Framework}
\label{sec:rl}

We adopt AWM (Advantage Weighted Matching)~\citep{awm2025} as the RL training algorithm. AWM is an RL optimization method specifically designed for diffusion models. Compared to SDE-based methods such as Flow-GRPO~\citep{liu2025flowgrpo}, AWM uses pure ODE sampling, avoiding image quality degradation caused by exploration noise. The overall training pipeline is illustrated in Figure~\ref{fig:pipeline}: sample prompts from the caption dataset $\to$ T2I model generates a group of images $\to$ frozen VLM computes PromptEcho rewards $\to$ compute group-relative advantages $\to$ AWM updates T2I model parameters.

\subsection{Data Preparation}
\label{sec:data}

The training data construction process is simple and efficient. We construct an internal dataset of approximately 100K concept-rich, high-quality images, and use Qwen3-VL-32B to generate detailed captions for these images, using the VLM prompt ``Describe this image in detail.'' Qwen3-VL-32B typically generates detailed captions of 200--400 English words, covering objects, attributes, spatial relationships, colors, textures, and other multi-dimensional information in the images.

Notably, the VLM (Qwen3-VL-32B) and prompt (``Describe this image in detail.'') used during the data preparation stage are kept consistent with those used for computing PromptEcho rewards during subsequent RL training. This design choice is motivated by the fact that different VLMs and different queries/prompts lead to different prior distributions of generated captions; maintaining consistency ensures that the training data distribution matches the conditional distribution used for reward computation, enabling PromptEcho rewards to more accurately measure image-text alignment.

After construction, the subsequent RL training process only requires the caption text and no longer needs the original images. This property makes the data preparation cost extremely low, and the caption dataset can be conveniently expanded and updated.

\section{Experiments}
\label{sec:experiments}

\subsection{Experimental Setup}
\label{sec:setup}

\paragraph{Models and Training Configuration.} Our main experiments are conducted on two state-of-the-art open-source T2I models: Z-Image~\citep{cai2025zimage} and QwenImage-2512~\citep{wu2025qwenimage}. Unless otherwise stated, the VLM used for reward computation is a frozen Qwen3-VL-32B-Instruct~\citep{bai2025qwen3} (32 billion parameters), with no fine-tuning throughout training. Training uses LoRA (rank 64, alpha 128) fine-tuning with a learning rate of 5e-5. Each iteration samples 32 unique prompts, each generating 8 images (256 total), with an effective training batch size of 128. To improve sample efficiency, each batch of sampled data is trained for 4 epochs. On 32 H20 GPUs, Z-Image training takes approximately 100 hours and QwenImage-2512 approximately 200 hours. Both training and testing use a resolution of 1024$\times$1024. Training uses a CFG of 3.5 with 30-step ODE sampling; evaluation uses a CFG of 4.0 with 50 inference steps.

\paragraph{Evaluation Benchmarks.}

\textit{DenseAlignBench (ours).} To rigorously test the prompt following capability of state-of-the-art T2I models on concept-rich, detailed descriptions, we construct DenseAlignBench---a test set of dense captions with rich concepts, relationships, and fine-grained details. DenseAlignBench consists of 2000 test captions separated from our internal dataset; these captions are not in the training set but share the same source and distribution as the training data. We use Gemini-3-flash-preview~\citep{gemini3flash2025} for pairwise evaluation on the prompt following dimension, directly reflecting the method's effectiveness. We have open-sourced the DenseAlignBench\footnote{\url{https://github.com/roooobotx/prompt_echo}}.

Additionally, we evaluate the generalization of prompt following improvements on the following public benchmarks:

\textit{GenEval}~\citep{ghosh2023geneval}: a structured T2I evaluation benchmark covering single object, two objects, counting, colors, spatial position, and attribute binding sub-tasks.

\textit{DPG-Bench}~\citep{hu2024ella}: evaluates T2I models' semantic alignment with complex text descriptions using dense prompts, covering five categories (global, entity, attribute, relation, other), scored by the mPLUG-large VQA model.

\textit{TIIFBench}~\citep{wei2025tiifbench}: a fine-grained instruction following benchmark covering 40 dimensions, aggregated across three levels (Basic Following, Advanced Following, Real World Following), using GPT-4o as evaluator.

\subsection{Main Results}
\label{sec:main_results}

\subsubsection{DenseAlignBench}
\label{sec:vcg}

To directly validate the effectiveness of PromptEcho, we use captions from DenseAlignBench (sharing the same source and distribution as training data) to generate images with both the baseline models (Z-Image and QwenImage-2512) and the PromptEcho-trained models for the same prompts. We use Gemini-3-flash-preview for pairwise evaluation: each comparison randomly swaps image order with 50\% probability to eliminate position bias, with temperature set to 0.3 (detailed evaluation prompt in Appendix~\ref{app:vcg_prompt}). Results for both Z-Image and QwenImage-2512 are shown in Table~\ref{tab:vcg}.

\begin{table}[t]
  \caption{DenseAlignBench pairwise evaluation results. Pairwise comparison on the prompt following using Gemini-3-flash-preview.}
  \label{tab:vcg}
  \centering
  \resizebox{\linewidth}{!}{
  \begin{tabular}{lcccc}
    \toprule
    Model & Win Rate & Baseline Win Rate & Tie Rate & Net Advantage$\uparrow$ \\
    \midrule
    Z-Image: + PromptEcho vs Baseline & \textbf{61.5\%} & 34.7\% & 3.8\% & \textbf{+26.8pp} \\
    QwenImage-2512: + PromptEcho vs Baseline & \textbf{53.3\%} & 37.0\% & 9.7\% & \textbf{+16.2pp} \\
    \bottomrule
  \end{tabular}
  }
\end{table}

On Z-Image, the PromptEcho-trained model achieves an overwhelming advantage in prompt following, with a win rate of 61.5\%---nearly double the baseline (34.7\%)---and a net advantage of +26.8pp. On QwenImage-2512, it similarly achieves a significant win (53.3\% vs 37.0\%, net advantage +16.2pp), fully validating the effectiveness of PromptEcho as a reward signal and its cross-model generalization.

\subsubsection{Public Benchmark}
\label{sec:public_bench}

It is important to emphasize that the test prompts of the following public benchmarks differ significantly in distribution from our training data, which consists of approximately 200-word detailed captions generated by Qwen3-VL-32B. We perform no task-specific training for any of these benchmarks; the following results entirely reflect the generalized improvement in prompt following capability. Both Z-Image and QwenImage-2512 baselines use the officially released models, evaluated at 1024$\times$1024 resolution with CFG of 4.0 and 50 inference steps.

\paragraph{GenEval.}

\begin{table}[t]
  \caption{GenEval evaluation results. Evaluating T2I models' prompt following capability on structured sub-tasks.}
  \label{tab:geneval}
  \centering
  \resizebox{\linewidth}{!}{
  \begin{tabular}{lccccccc}
    \toprule
    Model & Single Obj. & Two Obj. & Counting & Colors & Position & Attr. Bind. & Overall$\uparrow$ \\
    \midrule
    Z-Image (Baseline) & 0.99 & 0.91 & 0.75 & 0.86 & 0.41 & 0.59 & 0.75 \\
    \textbf{+ PromptEcho} & \textbf{1.00} & \textbf{0.94} & \textbf{0.85} & 0.86 & \textbf{0.52} & \textbf{0.73} & \textbf{0.82} \\
    \midrule
    QwenImage-2512 (Baseline) & 0.99 & \textbf{0.94} & 0.56 & 0.87 & 0.47 & 0.64 & 0.74 \\
    \textbf{+ PromptEcho} & 0.99 & 0.93 & \textbf{0.68} & \textbf{0.90} & \textbf{0.55} & \textbf{0.70} & \textbf{0.79} \\
    \bottomrule
  \end{tabular}
  }
\end{table}

As shown in Table~\ref{tab:geneval}, on Z-Image, PromptEcho achieves improvements across all GenEval sub-tasks, with Overall increasing from 0.75 to 0.82 (+6.5pp). The most significant improvements include: Attribute Binding +14.0pp (0.59 $\to$ 0.73), indicating a qualitative leap in attribute binding capability; Position +11.0pp (0.41 $\to$ 0.52), substantially enhanced spatial relationship understanding; and Counting +10.0pp (0.75 $\to$ 0.85), significantly improved counting accuracy.

On QwenImage-2512, PromptEcho improves Overall from 0.74 to 0.79 (+5pp), with Counting (+12pp, 0.56 $\to$ 0.68), Position (+8pp, 0.47 $\to$ 0.55), and Attribute Binding (+6pp, 0.64 $\to$ 0.70) showing the most significant gains.

\paragraph{DPG-Bench.}

DPG-Bench evaluates T2I models' semantic alignment capability using dense prompts (long, complex descriptions), covering five categories: global, entity, attribute, relation, and other. Results are shown in Table~\ref{tab:dpgbench}.

\begin{table}[t]
  \caption{DPG-Bench evaluation results. Evaluating T2I models' semantic alignment with complex dense prompts.}
  \label{tab:dpgbench}
  \centering
  \resizebox{\linewidth}{!}{
  \begin{tabular}{lcccccc}
    \toprule
    Model & Global & Entity & Attribute & Relation & Other & Overall$\uparrow$ \\
    \midrule
    Z-Image (Baseline) & 91.60 & 91.54 & 90.32 & 92.76 & \textbf{91.94} & 86.90 \\
    \textbf{+ PromptEcho} & \textbf{93.05} & \textbf{92.76} & \textbf{91.87} & \textbf{93.89} & 89.99 & \textbf{87.92} \\
    \midrule
    QwenImage-2512 (Baseline) & \textbf{94.40} & \textbf{93.27} & 90.01 & 92.82 & 91.34 & 87.32 \\
    \textbf{+ PromptEcho} & 91.21 & 93.25 & \textbf{90.39} & \textbf{93.63} & \textbf{93.13} & \textbf{87.49} \\
    \bottomrule
  \end{tabular}
  }
\end{table}

On Z-Image, PromptEcho improves the DPG-Bench Overall from 86.90 to 87.92 (+1.02), with Attribute (+1.55) and Relation (+1.13) showing the most significant gains.

\paragraph{TIIFBench.}

TIIFBench evaluates T2I models' instruction following capability across multiple fine-grained dimensions, providing both short description and long description prompt forms. Results are shown in Table~\ref{tab:tiifbench}.

\begin{table}[t]
  \caption{TIIFBench evaluation results. Each cell contains two values corresponding to short/long description prompts.}
  \label{tab:tiifbench}
  \centering
  \resizebox{\linewidth}{!}{
  \begin{tabular}{l cc cccc cccccc c}
    \toprule
    \multirow{2}{*}{Model} & \multicolumn{2}{c}{Overall$\uparrow$} & \multicolumn{4}{c}{Basic Following} & \multicolumn{6}{c}{Advanced Following} & Real \\
    \cmidrule(lr){2-3} \cmidrule(lr){4-7} \cmidrule(lr){8-13} \cmidrule(lr){14-14}
     & short & long & Attr. & Rela. & Reas. & Avg & Attr.+Rela. & Attr.+Reas. & Rela.+Reas. & Style & Text & Avg & World \\
    \midrule
    Z-Image (Baseline) & 84.91 & 83.16 & 89.5/88.9 & 90.3/82.3 & 79.3/86.0 & 86.4/85.7 & 80.8/81.3 & 75.8/77.4 & 80.9/79.3 & 82.8/73.3 & 93.2/89.6 & 79.9/79.5 & 91.6/90.3 \\
    \textbf{+ PromptEcho} & \textbf{88.50} & \textbf{88.94} & \textbf{94.5/93.0} & \textbf{90.9/86.5} & \textbf{85.5/89.6} & \textbf{90.3/89.7} & \textbf{83.5/88.5} & \textbf{80.1/88.4} & \textbf{84.4/82.8} & \textbf{83.3/83.3} & \textbf{99.1/95.0} & \textbf{83.3/86.8} & \textbf{95.2/93.3} \\
    \midrule
    QwenImage-2512 (Baseline) & 84.89 & 83.25 & \textbf{92.2}/86.5 & \textbf{84.9}/86.8 & 79.4/81.0 & 85.5/84.8 & \textbf{84.5}/84.8 & 73.1/81.7 & \textbf{81.6}/81.6 & \textbf{80.0}/60.0 & 95.5/93.2 & 80.3/82.2 & 92.9/93.7 \\
    \textbf{+ PromptEcho} & \textbf{85.50} & \textbf{86.46} & 91.0/\textbf{87.0} & 84.7/\textbf{90.0} & \textbf{85.5}/\textbf{84.6} & \textbf{87.1}/\textbf{87.2} & 79.1/\textbf{85.4} & \textbf{80.2}/\textbf{88.6} & 81.3/\textbf{83.1} & 72.4/\textbf{69.0} & \textbf{99.1}/\textbf{95.4} & \textbf{80.6}/\textbf{85.4} & \textbf{96.2}/\textbf{95.0} \\
    \bottomrule
  \end{tabular}
  }
\end{table}

On Z-Image, PromptEcho achieves significant improvements across all levels of TIIFBench, with Overall improving by +3.6pp (short) / +5.8pp (long). At the sub-dimension level, Text (93.2$\to$99.1 short), Style (73.3$\to$83.3 long), and Real World (91.6$\to$95.2 short) show particularly notable gains. On QwenImage-2512, consistent improvements are also observed, with Overall +0.6pp (short) / +3.2pp (long), and significant gains in Reasoning (+6.1pp short), Text (+3.6pp short), and Real World (+3.3pp short).

The results across all public benchmarks (GenEval, DPG-Bench, TIIFBench) consistently demonstrate that, despite significant distributional differences between training and test data, the PromptEcho-trained models achieve substantial improvements on two T2I models with different architectures. We attribute this strong generalization to the nature of the PromptEcho reward signal---it originates from the image-text alignment knowledge of a large-scale VLM pretrained on massive data, rather than distribution-specific preference annotation data, thus naturally possessing cross-distribution, cross-architecture generalization capability.

\subsection{Ablation and Comparative Experiments}
\label{sec:ablation}

\subsubsection{PromptEcho vs InferScore}
\label{sec:ablation_inferscore}

To directly compare PromptEcho (computing continuous log-likelihood via the pretraining loss) with InferScore (generating discrete scores via autoregressive inference), we design a controlled ablation: both use the identical VLM (Qwen3-VL-32B) and the same base model (Z-Image), differing only in reward construction. Results are shown in Tables~\ref{tab:ablation_consolidated}--\ref{tab:ablation_vcg}.

\begin{table}[t]
  \caption{Ablation: PromptEcho vs InferScore on public benchmarks (Z-Image + Qwen3-VL-32B). GenEval reports per-subtask and overall scores; DPG-Bench reports per-category and overall scores; TIIFBench reports overall short/long scores.}
  \label{tab:ablation_consolidated}
  \centering
  \resizebox{\linewidth}{!}{
  \begin{tabular}{l ccccccc cccccc cc}
    \toprule
    \multirow{2}{*}{Model} & \multicolumn{7}{c}{GenEval} & \multicolumn{6}{c}{DPG-Bench} & \multicolumn{2}{c}{TIIFBench} \\
    \cmidrule(lr){2-8} \cmidrule(lr){9-14} \cmidrule(lr){15-16}
     & Single & Two & Count & Color & Pos. & Attr. & Overall$\uparrow$ & Global & Entity & Attr. & Rel. & Other & Overall$\uparrow$ & Short$\uparrow$ & Long$\uparrow$ \\
    \midrule
    Baseline & 0.99 & 0.91 & 0.75 & 0.86 & 0.41 & 0.59 & 0.75 & 91.60 & 91.54 & 90.32 & 92.76 & \textbf{91.94} & 86.90 & 84.91 & 83.16 \\
    \textbf{+ PromptEcho} & \textbf{1.00} & \textbf{0.94} & \textbf{0.85} & 0.86 & \textbf{0.52} & \textbf{0.73} & \textbf{0.82} & \textbf{93.05} & 92.76 & \textbf{91.87} & \textbf{93.89} & 89.99 & \textbf{87.92} & \textbf{88.50} & \textbf{88.94} \\
    + InferScore & \textbf{1.00} & 0.91 & 0.79 & 0.86 & 0.41 & 0.59 & 0.76 & 78.74 & \textbf{93.17} & 90.21 & 92.56 & 82.49 & 86.86 & 82.40 & 84.64 \\
    \bottomrule
  \end{tabular}
  }
\end{table}

\begin{table}[t]
  \caption{Ablation: PromptEcho vs InferScore on DenseAlignBench (Z-Image + Qwen3-VL-32B).}
  \label{tab:ablation_vcg}
  \centering
  \resizebox{\linewidth}{!}{
  \begin{tabular}{lcccc}
    \toprule
    Model & Win Rate & Baseline Win Rate & Tie Rate & Net Advantage$\uparrow$ \\
    \midrule
    \textbf{+ PromptEcho} vs Baseline & \textbf{61.5\%} & 34.7\% & 3.8\% & \textbf{+26.8pp} \\
    + InferScore vs Baseline & 28.0\% & \textbf{31.3\%} & 40.8\% & \textbf{$-$3.3pp} \\
    \bottomrule
  \end{tabular}
  }
\end{table}

The results reveal a stark contrast. On DenseAlignBench, PromptEcho achieves +26.8pp net advantage while InferScore falls below the baseline ($-$3.3pp). On public benchmarks, PromptEcho consistently outperforms the baseline (GenEval +6.5pp, DPG-Bench +1.02, TIIFBench +3.6/+5.8pp), while InferScore shows negligible improvement or degradation (GenEval +1pp, DPG-Bench $-$0.04, TIIFBench short $-$2.5pp). The core reason for InferScore's failure is that its discrete scoring mechanism prevents the VLM from distinguishing subtle differences in prompt following across multiple images generated from the same prompt with different seeds---frequently assigning identical scores to all images, effectively nullifying the reward signal. In contrast, PromptEcho outputs continuous log-likelihood values that naturally capture fine-grained distinctions (see Section~\ref{sec:understanding}).

\subsubsection{Reward Model Size Ablation}
\label{sec:ablation_reward_size}

To investigate the effect of VLM size on PromptEcho's reward quality, we compare training results on Z-Image using Qwen3-VL-32B (32 billion parameters) and Qwen3-VL-8B (8 billion parameters) as the reward model. Both use identical training data and hyperparameters.

On DenseAlignBench (Table~\ref{tab:ablation_reward_size_internal}), both the 32B and 8B reward models yield significant win rate improvements, with the 32B's net advantage (+26.8pp) substantially higher than the 8B (+18.3pp), indicating that larger VLMs provide higher-quality reward signals.

\begin{table}[t]
  \caption{DenseAlignBench reward model size ablation: 32B vs 8B VLM, both based on Z-Image.}
  \label{tab:ablation_reward_size_internal}
  \centering
  \resizebox{\linewidth}{!}{
  \begin{tabular}{lcccc}
    \toprule
    Model & Win Rate & Baseline Win Rate & Tie Rate & Net Advantage$\uparrow$ \\
    \midrule
    \textbf{+ PromptEcho (32B Reward)} vs Baseline & \textbf{61.5\%} & 34.7\% & 3.8\% & \textbf{+26.8pp} \\
    + PromptEcho (8B Reward) vs Baseline & 57.6\% & 39.3\% & 3.2\% & +18.3pp \\
    \bottomrule
  \end{tabular}
  }
\end{table}

The generalization results on public benchmarks (Table~\ref{tab:ablation_reward_size}) further corroborate this trend.

\begin{table}[t]
  \caption{Public benchmark reward model size ablation: 32B vs 8B VLM, both based on Z-Image.}
  \label{tab:ablation_reward_size}
  \centering
  \begin{tabular}{lcccc}
    \toprule
    Model & GenEval$\uparrow$ & DPG$\uparrow$ & TIIF-S$\uparrow$ & TIIF-L$\uparrow$ \\
    \midrule
    Z-Image (Baseline) & 0.75 & 86.90 & 84.91 & 83.16 \\
    \textbf{+ PromptEcho (32B Reward)} & \textbf{0.82} & \textbf{87.92} & 88.50 & \textbf{88.94} \\
    + PromptEcho (8B Reward) & 0.77 & 87.15 & \textbf{89.08} & 87.69 \\
    \bottomrule
  \end{tabular}
\end{table}

The 32B reward outperforms the 8B reward on 3 out of 4 public metrics: GenEval (0.82 vs 0.77, +5pp), DPG-Bench (87.92 vs 87.15), and TIIF-long (88.94 vs 87.69). The gap is most pronounced on GenEval, indicating that the larger VLM provides more precise image-text alignment reward signals. This result demonstrates that PromptEcho's reward quality scales with VLM parameter count, further supporting the advantage of ``improving reward quality as VLMs advance.''

\subsection{Text Rendering Experiment: Validating the General-Purpose Reward Paradigm}
\label{sec:text_rendering}

To validate PromptEcho's potential as a general-purpose reward paradigm, we apply it to a task fundamentally different from general prompt following---optimizing text rendering quality in e-commerce posters. On an internal poster generation model, we extend PromptEcho to the text rendering scenario by adapting the VLM prompt and label format.

\paragraph{Method Adaptation.} Unlike the general prompt following experiments that use ``Describe this image in detail.'' as the VLM query, in the text rendering scenario we replace the VLM query with a structured OCR recognition prompt---requiring the VLM to recognize all designed marketing text in the image and classify it by semantic role into main\_title, subtitle, selling\_points, and other\_text. Correspondingly, the label is replaced from free-form captions to structured text labels (in JSON format) pre-extracted from editing instructions. The core computation of the PromptEcho reward (negative of the VLM cross-entropy loss) remains unchanged. Detailed prompt templates are provided in Appendix~\ref{app:text_render}.

\paragraph{Experimental Results.} Using Qwen3.5-Plus~\citep{qwen35plus2025} as an independent evaluation model---chosen for its strong Chinese character recognition capability, which is critical for accurate OCR-based evaluation in this scenario---we perform character-by-character strict comparison evaluation on 5000 test samples (details in Appendix~\ref{app:text_render_eval}). Here, text rendering accuracy refers to the proportion of images where all text in the entire image is completely correct (1 point for fully correct, 0 otherwise). After PromptEcho training, the model's text rendering accuracy improves from 68\% to 75\% (+7pp).

\paragraph{Significance for Generality.} This experiment demonstrates that PromptEcho's core idea---using VLM cross-entropy loss as the reward---possesses good generality. By adapting the VLM query and label format, PromptEcho can be flexibly applied to different T2I models and different optimization objectives, without retraining task-specific reward models. This further confirms the value of PromptEcho as a general-purpose reward construction paradigm.

\subsection{Qualitative Analysis}
\label{sec:qualitative}

We present qualitative improvements from PromptEcho training along two dimensions: general prompt following capability and text rendering accuracy.

\subsubsection{Improved General Prompt Following}

We select 6 representative cases from the DenseAlignBench pairwise evaluation (Appendix~\ref{app:qualitative_cases}), comparing QwenImage-2512 baseline with the PromptEcho-trained model on long, detailed text descriptions. We observe that these cases span diverse scenario types---character portraits, animal close-ups, still life arrangements, natural macro photography, and spatial relationships---and PromptEcho consistently demonstrates higher prompt fidelity in attribute details (color, quantity, texture), object features, and pose descriptions.

\subsubsection{Text Rendering Before and After Training}

Beyond general prompt following, we observe notable qualitative improvements from PromptEcho in the text rendering task (Section~\ref{sec:text_rendering}). Before training, the model frequently produces e-commerce posters with missing strokes, extra strokes, uneven character spacing, and incorrect Chinese characters. After PromptEcho training, the model shows clear improvements in stroke completeness, character accuracy, and overall layout regularity. Representative cases are shown in Appendix Figure~\ref{fig:text_rendering_cases}. This further validates PromptEcho's flexibility as a general-purpose reward paradigm---by simply adapting the VLM query and label format, the same reward mechanism can be transferred to fundamentally different optimization objectives.

\section{Discussion}
\label{sec:discussion}

\subsection{Pretraining Loss vs.\ Autoregressive Scoring}
\label{sec:understanding}

InferScore lets the VLM explicitly score images through autoregressive generation, a process subject to hallucinations and sampling randomness. More critically, its discrete scoring mechanism prevents the VLM from distinguishing subtle differences in prompt following across multiple images generated from the same prompt with different seeds---frequently assigning identical scores to all images, effectively nullifying the reward signal. PromptEcho instead directly computes continuous log-likelihood values via the VLM's pretraining cross-entropy loss, requiring no autoregressive generation. The result is fully deterministic and naturally captures fine-grained distinctions---as confirmed by the large performance gap in our ablation (Section~\ref{sec:ablation_inferscore}).

\subsection{Open-Source VLMs as Free Reward Annotators}

PromptEcho transforms the collective progress of the open-source VLM community into free reward signals for T2I optimization. Traditional reward models require dedicated annotation and specialized training, while PromptEcho directly leverages existing VLMs. As these VLMs improve, PromptEcho's reward quality improves in tandem---our 32B vs.\ 8B ablation (Section~\ref{sec:ablation_reward_size}) provides direct evidence of this scaling property.

\subsection{Limitations}

Our method has certain limitations. First, in domains where the VLM lacks specific visual expertise (e.g., fine-grained photographic aesthetics, recognition of subtle text errors), PromptEcho may not provide effective reward signals. Second, since PromptEcho's reward computation is based on the VLM's forward pass rather than autoregressive reasoning, it does not leverage the more powerful reasoning and chain-of-thought capabilities of state-of-the-art large models, which may limit the precision of reward signals in scenarios requiring complex reasoning and judgment.

\section{Conclusion}
\label{sec:conclusion}

We presented PromptEcho, which directly uses the cross-entropy loss of a frozen VLM as an annotation-free, training-free reward signal for T2I reinforcement learning. The key insight is that the VLM's pretraining loss already encodes rich image-text alignment knowledge that can be extracted through a single deterministic forward pass. Experiments on Z-Image and QwenImage-2512 demonstrate substantial improvements on DenseAlignBench (+26.8pp / +16.2pp) and consistent gains across GenEval, DPG-Bench, and TIIFBench. Ablation studies confirm that PromptEcho far outperforms inference-based scoring and that reward quality scales with VLM size. Successful application to text rendering further validates PromptEcho as a general-purpose reward paradigm.

Current limitations include reduced effectiveness in domains where the VLM lacks visual expertise, and the inability to leverage chain-of-thought reasoning. Future directions include scaling to larger VLMs and training data, domain-specific VLM adaptation, and extension to text-to-video generation.

\appendix

\section{DenseAlignBench Pairwise Evaluation Prompt Template}
\label{app:vcg_prompt}

The following prompt template is used in the DenseAlignBench evaluation, where Gemini-3-flash-preview performs pairwise comparison on the prompt following for each pair of images:

{\small
\begin{verbatim}
You are a professional image quality assessment expert specializing in
evaluating prompt following accuracy.

Your task is to compare two AI-generated images and determine which
image better follows the given prompt.
Focus only on prompt following accuracy - do not consider aesthetics,
artistic quality, or realism.

**Prompt:**
{prompt}

**Evaluation Process:**
1. Carefully read and understand all requirements in the prompt:
   - Main subjects and objects
   - Actions and poses
   - Visual attributes (color, size, material, texture)
   - Composition and layout
   - Style and atmosphere
   - Any text or written elements
   - Spatial relationships (foreground, background, position)
   - Quantities and counting

2. Examine Image A:
   - Which prompt requirements are accurately depicted?
   - Which prompt requirements are missing or incorrect?
   - Are there elements not mentioned in the prompt?

3. Examine Image B:
   - Which prompt requirements are accurately depicted?
   - Which prompt requirements are missing or incorrect?
   - Are there elements not mentioned in the prompt?

4. Compare:
   - Which image more accurately captures the prompt requirements?
   - If both are equal in accuracy, select "tie"

**Preference Options:**
- "image_a": Image A better follows the prompt than Image B
- "image_b": Image B better follows the prompt than Image A
- "tie": Both images follow the prompt to a similar degree

**Important Notes:**
- Make your decision based solely on prompt following accuracy
- Ignore differences in artistic style, aesthetics, or realism
- Remain objective and unbiased
- Select "tie" when prompt following quality is truly similar

**Output Format (strict JSON, no markdown):**
{
  "reasoning": "<detailed explanation>",
  "preference": "<image_a, image_b, or tie>"
}
\end{verbatim}
}

\section{Text Rendering Experiment Prompt Templates}
\label{app:text_render}

\subsection{Label Construction Prompt}

Used to extract structured text labels from editing instructions, serving as the label for PromptEcho reward computation:

{\small
\begin{verbatim}
You are a professional text structure analysis expert. Your task is to
extract all text content required to appear in the image from a given
prompt, and classify them by semantic role in a structured manner.

**Given Prompt:**
{prompt}

**Extraction Rules:**

1. **main_title**: If the title has multiple lines (e.g., top line/
   bottom line), each line is a separate element in the list.
2. **subtitle**: Text inferred to be subtitles, such as product
   titles or secondary headings.
3. **selling_points**: Product selling points, recommendations,
   or introductory phrases.
4. **other_text**: Other text not belonging to the above categories.

**Notes:**
- Each piece of text should be extracted completely, preserving
  the original content
- If a category has no corresponding content, return an empty list []
- Carefully read the context to determine the semantic role of
  each piece of text

**Output Format (must strictly follow JSON format, do not add any
markdown tags):**
{
  "reasoning": "<brief explanation of how you identified and
   classified these texts>",
  "main_title": ["first line title", "second line title"],
  "subtitle": ["subtitle text"],
  "selling_points": ["point 1", "point 2"],
  "other_text": ["other text 1", "other text 2"]
}
\end{verbatim}
}

\subsection{Training Reward Computation Prompt}

After inputting the generated image, the OCR recognition prompt used as the VLM query (with structured text labels as the label for cross-entropy computation):

{\small
\begin{verbatim}
You are a professional image text recognition expert. Your task is to
recognize all designed and typeset text content in this e-commerce
marketing image, and classify them by semantic role in a structured
manner.

**Extraction Rules:**

1. **main_title**: The largest and most prominent title text in the
   image. If the title has multiple lines, each line is a separate
   element in the list.
2. **subtitle**: Subtitles or supplementary text, such as product
   titles or secondary headings.
3. **selling_points**: Product selling points, recommendations,
   introductory phrases, efficacy features, etc.
4. **other_text**: Other text not belonging to the above categories.

**Notes:**
- Only extract designed marketing text in the image; do not extract
  text printed on product packaging or physical objects
- Each piece of text should be extracted completely, preserving
  the original content
- Each field is a list of strings, arranged in visual reading order
- If a category has no corresponding content, return an empty list []

**Output Format (must strictly follow JSON format, do not add any
markdown tags or explanations):**
{"main_title": ["first line title", "second line title"],
 "subtitle": ["subtitle text"],
 "selling_points": ["point 1", "point 2"],
 "other_text": ["other text 1", "other text 2"]}
\end{verbatim}
}

\subsection{Text Rendering Evaluation Prompt}
\label{app:text_render_eval}

The prompt used for independent evaluation, where the VLM performs character-by-character strict comparison and outputs a binary score (0 or 1):

{\small
\begin{verbatim}
You are a professional and strict text rendering quality inspector.
You are extremely sensitive to text details - any character error,
garbled text, missing character, extra character, or positional
deviation is unacceptable.

Your task is to check whether the text rendered in the image exactly
matches the text required by the prompt.

**Prompt:**
{prompt}

**Inspection Steps:**

1. **Extract required text**: Carefully read the prompt and extract
   all text content required to appear in the image. Pay attention
   to distinguishing Chinese/English, uppercase/lowercase, numbers,
   etc. If the prompt does not explicitly require any text to appear
   in the image, expected_text is an empty string, and directly
   give 1 point.

2. **Recognize image text**: Carefully examine the image and
   recognize all text appearing in it character by character.
   Ignore text on product packaging; focus on the text required
   by the prompt.

3. **Character-by-character comparison**: Perform strict
   character-by-character comparison between the prompt-required
   text and the actual text in the image:
   - Every required character must be completely correct;
     garbled text is unacceptable
   - No extra characters, missing characters, wrong characters,
     or garbled text allowed
   - No duplicate text (unless the prompt explicitly requires it)
   - Text order must be correct

4. **Scoring**:
   - 1 point: All required text is rendered completely correctly
     in the image with no character-level errors
   - 0 points: Any text error exists

**Output Format (strict JSON):**
{
  "expected_text": "<all required text extracted from the prompt>",
  "found_text": "<all text actually recognized from the image>",
  "reasoning": "<detailed character-by-character comparison results>",
  "score": <integer 0 or 1>
}
\end{verbatim}
}

\section{Qualitative Analysis Cases}
\label{app:qualitative_cases}

\begin{figure}[htbp]
\centering
\includegraphics[width=0.9\textwidth]{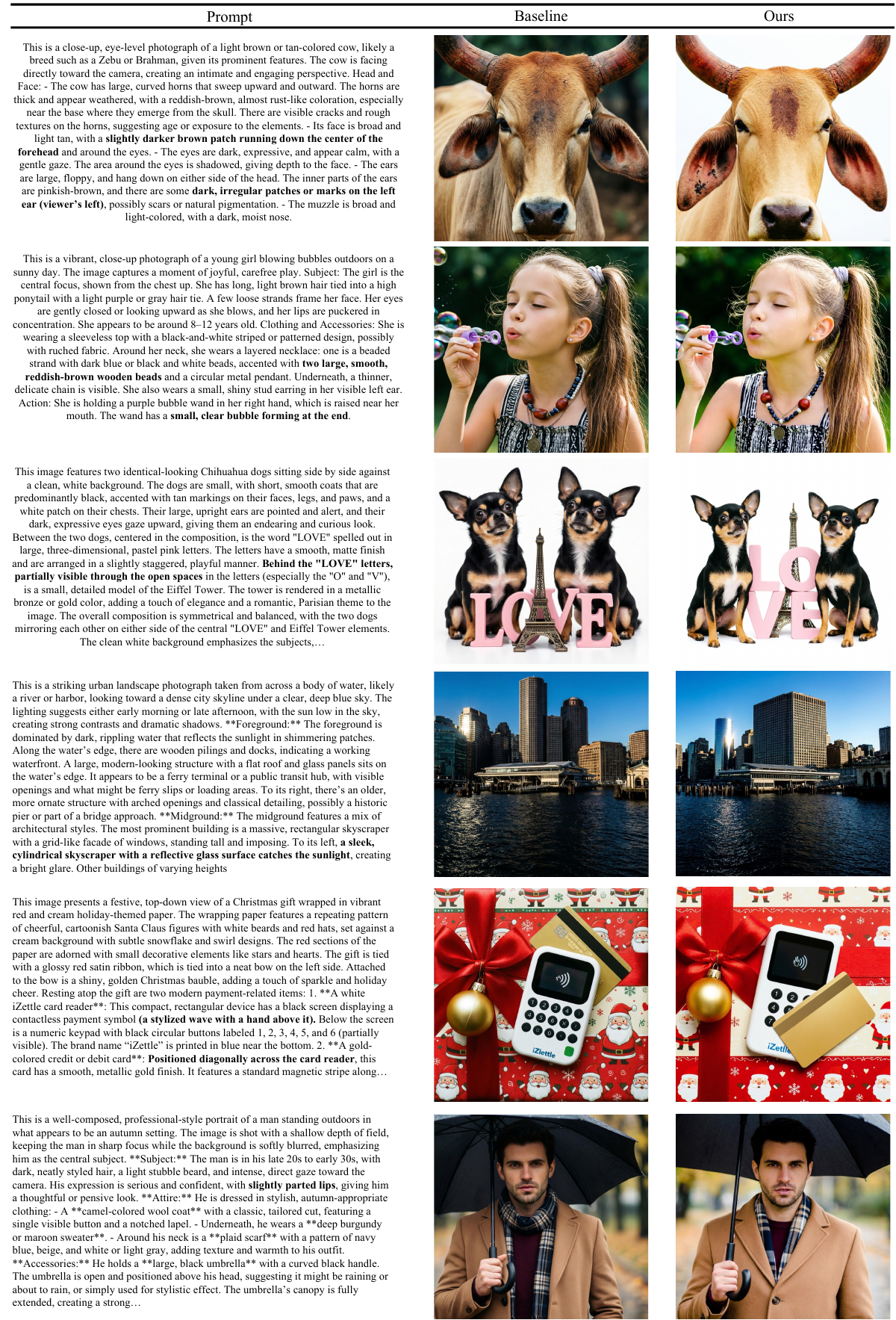}
\caption{Qualitative comparison of 6 representative winning cases. Each row shows the full \textbf{Prompt}, \textbf{Baseline} (QwenImage-2512), and \textbf{Ours} (PromptEcho-trained). Ours demonstrates more faithful adherence to fine-grained details including color, count, texture, and spatial relations.}
\label{fig:qualitative_cases}
\end{figure}

\begin{figure}[htbp]
\centering
\includegraphics[width=0.9\textwidth]{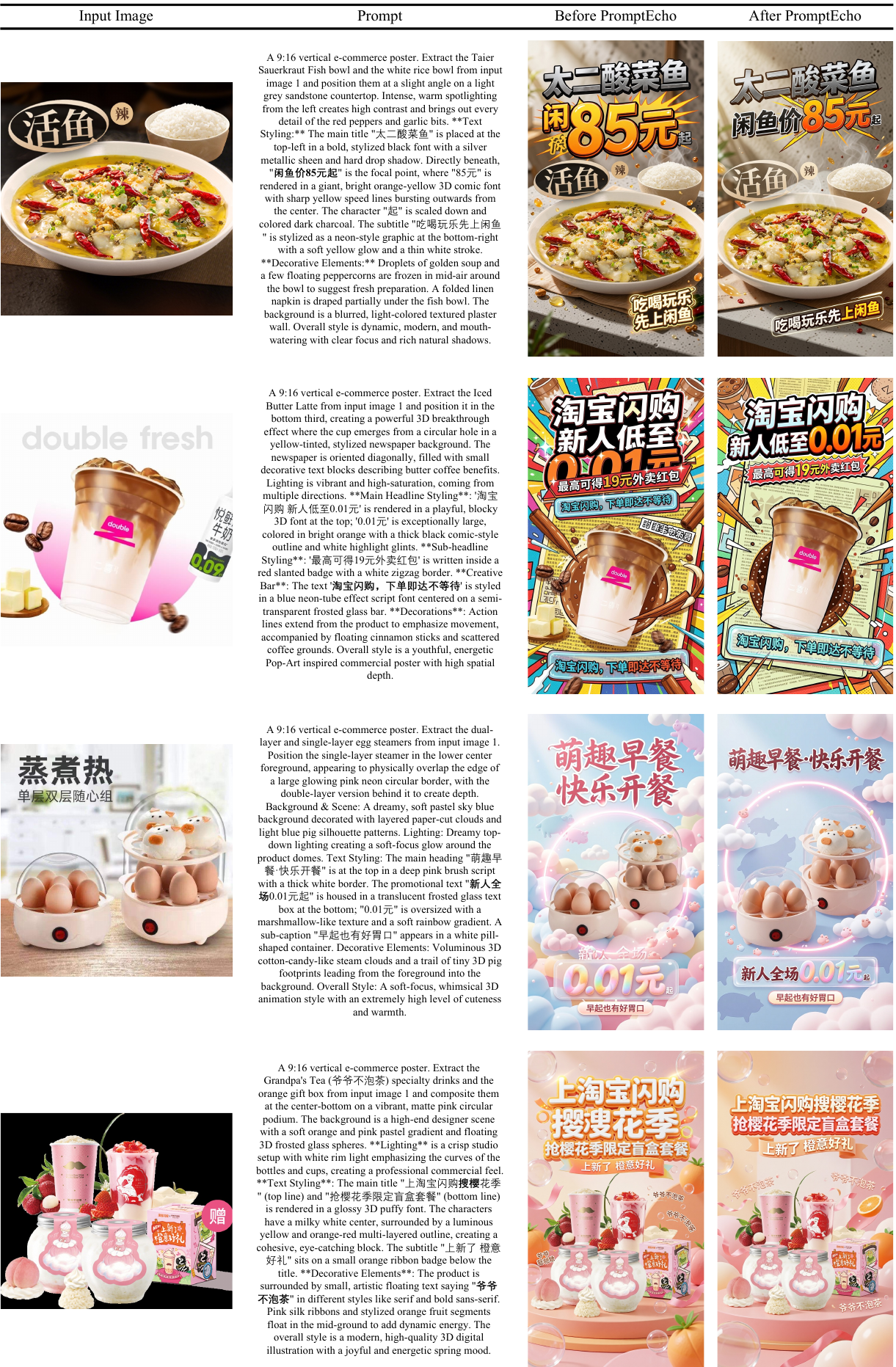}
\caption{Text rendering quality comparison before and after PromptEcho training (Section~\ref{sec:text_rendering}). Each row shows the \textbf{Input Image}, \textbf{Prompt}, \textbf{Before PromptEcho}, and \textbf{After PromptEcho}. After training, the model produces more accurate Chinese characters with improved stroke completeness and layout consistency in e-commerce poster generation.}
\label{fig:text_rendering_cases}
\end{figure}

\typeout{get arXiv to do 4 passes: Label(s) may have changed. Rerun}

\begin{thebibliography}{24}
\providecommand{\natexlab}[1]{#1}
\providecommand{\url}[1]{\texttt{#1}}
\expandafter\ifx\csname urlstyle\endcsname\relax
  \providecommand{\doi}[1]{doi: #1}\else
  \providecommand{\doi}{doi: \begingroup \urlstyle{rm}\Url}\fi

\bibitem[Black et~al.(2024)Black, Janner, Du, Kostrikov, and
  Levine]{black2024ddpo}
Kevin Black, Michael Janner, Yilun Du, Ilya Kostrikov, and Sergey Levine.
\newblock Training diffusion models with reinforcement learning.
\newblock In \emph{International Conference on Learning Representations
  (ICLR)}, 2024.

\bibitem[Fan et~al.(2023)Fan, Watkins, Du, Liu, Ryu, Boutilier, Abbeel,
  Ghavamzadeh, Lee, and Lee]{fan2024dpok}
Ying Fan, Olivia Watkins, Yuqing Du, Hao Liu, Moonkyung Ryu, Craig Boutilier,
  Pieter Abbeel, Mohammad Ghavamzadeh, Kangwook Lee, and Kimin Lee.
\newblock {DPOK}: Reinforcement learning for fine-tuning text-to-image
  diffusion models.
\newblock In \emph{Advances in Neural Information Processing Systems
  (NeurIPS)}, 2023.

\bibitem[Liu et~al.(2025)Liu, Liu, Liang, Li, Liu, Wang, Wan, Zhang, and
  Ouyang]{liu2025flowgrpo}
Jie Liu, Gongye Liu, Jiajun Liang, Yangguang Li, Jiaheng Liu, Xintao Wang,
  Pengfei Wan, Di~Zhang, and Wanli Ouyang.
\newblock Flow-{GRPO}: Training flow matching models via online {RL}.
\newblock \emph{arXiv preprint arXiv:2505.05470}, 2025.

\bibitem[Xue et~al.(2025)Xue, Ge, Zhang, Li, and Ma]{awm2025}
Shuchen Xue, Chongjian Ge, Shilong Zhang, Yichen Li, and Zhi-Ming Ma.
\newblock Advantage weighted matching: Aligning {RL} with pretraining in
  diffusion models.
\newblock \emph{arXiv preprint arXiv:2509.25050}, 2025.

\bibitem[Hessel et~al.(2021)Hessel, Holtzman, Forbes, Le~Bras, and
  Choi]{hessel2021clipscore}
Jack Hessel, Ari Holtzman, Maxwell Forbes, Ronan Le~Bras, and Yejin Choi.
\newblock {CLIPScore}: A reference-free evaluation metric for image captioning.
\newblock In \emph{Proceedings of the 2021 Conference on Empirical Methods in
  Natural Language Processing (EMNLP)}, 2021.

\bibitem[Kirstain et~al.(2023)Kirstain, Polyak, Singer, Matiana, Penna, and
  Levy]{kirstain2023pickapic}
Yuval Kirstain, Adam Polyak, Uriel Singer, Shahbuland Matiana, Joe Penna, and
  Omer Levy.
\newblock Pick-a-pic: An open dataset of user preferences for text-to-image
  generation.
\newblock In \emph{Advances in Neural Information Processing Systems
  (NeurIPS)}, 2023.

\bibitem[Xu et~al.(2023)Xu, Liu, Wu, Tong, Li, Ding, Tang, and
  Dong]{xu2023imagereward}
Jiazheng Xu, Xiao Liu, Yuchen Wu, Yuxuan Tong, Qinkai Li, Ming Ding, Jie Tang,
  and Yuxiao Dong.
\newblock {ImageReward}: Learning and evaluating human preferences for
  text-to-image generation.
\newblock In \emph{Advances in Neural Information Processing Systems
  (NeurIPS)}, 2023.

\bibitem[Wu et~al.(2023)Wu, Hao, Sun, Chen, Zhu, Zhao, and Li]{wu2023hpsv2}
Xiaoshi Wu, Yiming Hao, Keqiang Sun, Yixiong Chen, Feng Zhu, Rui Zhao, and
  Hongsheng Li.
\newblock Human preference score v2: A solid benchmark for evaluating human
  preferences of text-to-image synthesis.
\newblock \emph{arXiv preprint arXiv:2306.09341}, 2023.

\bibitem[Xu et~al.(2024)Xu, Huang, Cheng, et~al.]{xu2024visionreward}
Jiazheng Xu, Yu~Huang, Jiale Cheng, et~al.
\newblock {VisionReward}: Fine-grained multi-dimensional human preference
  learning for image and video generation.
\newblock \emph{arXiv preprint arXiv:2412.21059}, 2024.

\bibitem[Wang et~al.(2025)Wang, Zang, Li, Jin, and Wang]{wang2025unifiedreward}
Yibin Wang, Yuhang Zang, Hao Li, Cheng Jin, and Jiaqi Wang.
\newblock Unified reward model for multimodal understanding and generation.
\newblock \emph{arXiv preprint arXiv:2503.05236}, 2025.

\bibitem[Wu et~al.(2025{\natexlab{a}})Wu, Gao, Ye, Li, Li, Guo, Liu, Xue, Hou,
  Liu, et~al.]{rewarddance2025}
Jie Wu, Yu~Gao, Zilyu Ye, Ming Li, Liang Li, Hanzhong Guo, Jie Liu, Zeyue Xue,
  Xiaoxia Hou, Wei Liu, et~al.
\newblock {RewardDance}: Reward scaling in visual generation.
\newblock \emph{arXiv preprint arXiv:2509.08826}, 2025{\natexlab{a}}.

\bibitem[Wallace et~al.(2024)Wallace, Dang, Rafailov, Zhou, Lou, Purushwalkam,
  Ermon, Xiong, Joty, and Naik]{wallace2024diffusiondpo}
Bram Wallace, Meihua Dang, Rafael Rafailov, Linqi Zhou, Aaron Lou, Senthil
  Purushwalkam, Stefano Ermon, Caiming Xiong, Shafiq Joty, and Nikhil Naik.
\newblock Diffusion model alignment using direct preference optimization.
\newblock In \emph{IEEE/CVF Conference on Computer Vision and Pattern
  Recognition (CVPR)}, 2024.

\bibitem[Gu et~al.(2024)Gu, Wang, Yin, Xie, and Zhou]{gu2024diffusionrpo}
Yi~Gu, Zhendong Wang, Yueqin Yin, Yujia Xie, and Mingyuan Zhou.
\newblock Diffusion-{RPO}: Aligning diffusion models through relative
  preference optimization.
\newblock \emph{arXiv preprint arXiv:2406.06382}, 2024.

\bibitem[Liu et~al.(2023)Liu, Li, Wu, and Lee]{liu2024llava}
Haotian Liu, Chunyuan Li, Qingyang Wu, and Yong~Jae Lee.
\newblock Visual instruction tuning.
\newblock In \emph{Advances in Neural Information Processing Systems
  (NeurIPS)}, 2023.

\bibitem[Chen et~al.(2023)Chen, Wu, Wang, Su, Chen, Xing, Zhong, Zhang, Zhu,
  Lu, et~al.]{chen2024internvl}
Zhe Chen, Jiannan Wu, Wenhai Wang, Weijie Su, Guo Chen, Sen Xing, Muyan Zhong,
  Qinglong Zhang, Xizhou Zhu, Lewei Lu, et~al.
\newblock {InternVL}: Scaling up vision foundation models and aligning for
  generic visual-linguistic tasks.
\newblock \emph{arXiv preprint arXiv:2312.14238}, 2023.

\bibitem[Bai et~al.(2023)Bai, Bai, Yang, Wang, Tan, Wang, Lin, Zhou, and
  Zhou]{bai2023qwenvl}
Jinze Bai, Shuai Bai, Shusheng Yang, Shijie Wang, Sinan Tan, Peng Wang, Junyang
  Lin, Chang Zhou, and Jingren Zhou.
\newblock Qwen-{VL}: A versatile vision-language model for understanding,
  localization, text reading, and beyond.
\newblock \emph{arXiv preprint arXiv:2308.12966}, 2023.

\bibitem[Cai et~al.(2025)Cai, Cao, Du, Gao, Hoi, Hou, Huang, Jiang, Jin, Li,
  et~al.]{cai2025zimage}
Huanqia Cai, Sihan Cao, Ruoyi Du, Peng Gao, Steven Hoi, Zhaohui Hou, Shijie
  Huang, Dengyang Jiang, Xin Jin, Liangchen Li, et~al.
\newblock Z-image: An efficient image generation foundation model with
  single-stream diffusion transformer.
\newblock \emph{arXiv preprint arXiv:2511.22699}, 2025.

\bibitem[Wu et~al.(2025{\natexlab{b}})Wu, Li, Zhou, Lin, Gao, Yan, Yin, Bai,
  Xu, Chen, et~al.]{wu2025qwenimage}
Chenfei Wu, Jiahao Li, Jingren Zhou, Junyang Lin, Kaiyuan Gao, Kun Yan,
  Sheng-ming Yin, Shuai Bai, Xiao Xu, Yilei Chen, et~al.
\newblock Qwen-image technical report.
\newblock \emph{arXiv preprint arXiv:2508.02324}, 2025{\natexlab{b}}.

\bibitem[Bai et~al.(2025)Bai, Cai, Chen, Chen, Chen, Cheng, Deng, Ding, Gao,
  Ge, et~al.]{bai2025qwen3}
Shuai Bai, Yuxuan Cai, Ruizhe Chen, Keqin Chen, Xionghui Chen, Zesen Cheng,
  Lianghao Deng, Wei Ding, Chang Gao, Chunjiang Ge, et~al.
\newblock Qwen3-vl technical report.
\newblock \emph{arXiv preprint arXiv:2511.21631}, 2025.

\bibitem[{Google DeepMind}(2025)]{gemini3flash2025}
{Google DeepMind}.
\newblock Gemini 3 flash preview.
\newblock
  \url{https://ai.google.dev/gemini-api/docs/models/gemini-3-flash-preview},
  2025.

\bibitem[Ghosh et~al.(2023)Ghosh, Hajishirzi, and Schmidt]{ghosh2023geneval}
Dhruba Ghosh, Hanna Hajishirzi, and Ludwig Schmidt.
\newblock {GenEval}: An object-focused framework for evaluating text-to-image
  alignment.
\newblock \emph{arXiv preprint arXiv:2310.11513}, 2023.

\bibitem[Hu et~al.(2024)Hu, Wang, Fang, Fu, Cheng, and Yu]{hu2024ella}
Xiwei Hu, Rui Wang, Yixiao Fang, Bin Fu, Pei Cheng, and Gang Yu.
\newblock {ELLA}: Equip diffusion models with {LLM} for enhanced semantic
  alignment.
\newblock \emph{arXiv preprint arXiv:2403.05135}, 2024.

\bibitem[Wei et~al.(2025)Wei, Zhang, Wang, Wei, Guo, and
  Zhang]{wei2025tiifbench}
Xinyu Wei, Jinrui Zhang, Zeqing Wang, Hongyang Wei, Zhen Guo, and Lei Zhang.
\newblock {TIIF-Bench}: How does your {T2I} model follow your instructions?
\newblock \emph{arXiv preprint arXiv:2506.02161}, 2025.

\bibitem[{Qwen Team}(2025)]{qwen35plus2025}
{Qwen Team}.
\newblock Qwen3.5-397b-a17b.
\newblock \url{https://huggingface.co/Qwen/Qwen3.5-397B-A17B}, 2025.

\end{thebibliography}
\end{document}